%% file: main.tex
\documentclass[times,twocolumn,final,authoryear]{elsarticle}
\pdfoutput=1

\usepackage{ycviu}
\usepackage{framed,multirow}
\usepackage{amssymb}
\usepackage{latexsym}
\usepackage[utf8]{inputenc} %
\usepackage[T1]{fontenc}    %
\usepackage{hyperref}       %
\usepackage{url}            %
\usepackage{booktabs}       %
\usepackage{amsfonts}       %
\usepackage{nicefrac}       %
\usepackage{microtype}      %
\usepackage{booktabs}
\usepackage{tikz}
\usepackage{lipsum}
\usepackage{overpic}
\usepackage{amsmath}
\usepackage{xspace}
\usepackage{adjustbox}
\usepackage{todonotes}
\usepackage{etoolbox} %
\robustify\bfseries %
\usepackage{booktabs} %
\usepackage{array}
\usepackage{graphicx,subcaption} %
\usepackage[tight-spacing=true]{siunitx}
\usepackage{color,soul}
\usepackage{xcolor}
\usepackage{diagbox} %
\usepackage{cleveref}

\presetkeys%
    {todonotes}%
    {inline,backgroundcolor=yellow}{}

\input{macro}
\definecolor{newcolor}{rgb}{.8,.349,.1}
\makeatletter
\DeclareRobustCommand\onedot{\futurelet\@let@token\@onedot}
\def\@onedot{\ifx\@let@token.\else.\null\fi\xspace}
\makeatother

\journal{Computer Vision and Image Understanding}
\begin{document}
\clearpage

\ifpreprint
  \setcounter{page}{1}
\else
  \setcounter{page}{1}
\fi

\begin{frontmatter}
\title{\paperTitle}
\author[1]{Sebastien \snm{Ehrhardt}\corref{cor1}} 
\ead{hyenal@robots.ox.ac.uk}
\author[2]{Aron \snm{Monszpart}}
\ead{aron.monszpart.12@cs.ucl.uk}
\author[2]{Niloy J. \snm{Mitra}}
\ead{n.mitra@cs.ucl.ac.uk}
\author[1]{Andrea \snm{Vedaldi}}
\ead{vedaldi@robots.ox.ac.uk}

\address[1]{Department of Engineering Science, Parks road, Oxford, United Kingdom}
\address[2]{Department of Computer Science, Gower St, London, United Kingdom}

\received{December 2017}
\finalform{xxx}
\accepted{xxx}
\availableonline{xxx}
\communicated{S. Ehrhardt}

\begin{abstract}
\input{abstract}
\end{abstract}

\end{frontmatter}

\input{intro}

\input{relatedWork}

\input{method}

\input{experimentalSetup}

\input{results}

\input{conclusions}

\section*{Acknowledgments}
The authors would like to gratefully acknowledge the support of ERC 638009-IDIU and ERC SmartGeometry StG-2013-335373 grants.
\bibliographystyle{model2-names}
\bibliography{learning2Predict}
\end{document}

%% file: macro.tex
\crefname{figure}{Fig.}{Fig.}
\crefname{figure}{Fig.}{Fig.}
\crefname{table}{Table}{Table}

\newcommand{\paperTitle}{Taking Visual Motion Prediction To New Heightfields}

\newcommand{\ninputs}{T_0}

\newcommand{\bv}{\mathbf{v}}

\newcommand{\bx}{\mathbf{x}}

\newcommand{\revisionm}[1]{#1}
\newcommand{\revision}[1]{#1}

\newcommand{\TNet}{\textit{DispNet}\xspace} %
\newcommand{\PTNet}{\textit{PosNet}\xspace} %
\newcommand{\PNet}{\textit{ProbNet}\xspace} %
\newcommand{\IntNet}{\textit{InterpNet}\xspace} %

\newcommand{\noentry}{{\text{--}}}

\newcolumntype{C}[1]{>{\centering\let\newline\\\arraybackslash\hspace{0pt}}m{#1}}

\newcommand{\y}{\mathbf{y}}
\newcommand{\x}{\mathbf{x}}
\newcommand{\h}{\mathbf{h}}

\newcommand{\woa}{w/o ang. vel.}
\newcommand{\Ttrain}{$T_\text{train}$}
\newcommand{\Text}{$T_\text{gen}$}

\def\eg{\emph{e.g}\onedot} 
\def\ie{\emph{i.e}\onedot}

\def\etc{\emph{etc}\onedot}

\newcommand{\bowlone}{\text{Hemispherical bowl}\xspace}
\newcommand{\bowltwo}{\text{Ellipsoidal bowl}\xspace}
\newcommand{\bowloneq}{`\bowlone'\xspace}
\newcommand{\bowltwoq}{`\bowltwo'\xspace}

\newcommand{\pos}{\text{pixel}\xspace}
\newcommand{\av}{\text{ang. vel.}\xspace}

\newcommand*{\realdata}{}%

%% file: abstract.tex
While the basic laws of Newtonian mechanics are well understood, explaining a physical scenario still requires manually modeling the problem with suitable equations and estimating the associated parameters. 
In order to be able to leverage the approximation capabilities of artificial intelligence techniques in such physics related contexts, researchers have handcrafted relevant states, and then used neural networks to learn the state transitions using simulation runs as training data.
Unfortunately, such approaches are unsuited for modeling complex real-world scenarios, where manually authoring relevant state spaces tend to be tedious and challenging. 
In this work, we investigate if {\em neural networks can implicitly learn physical states of real-world mechanical processes only based on visual data} while internally modeling non-homogeneous environment 
and in the process enable long-term physical extrapolation. We develop a recurrent neural network architecture for this task and also characterize resultant uncertainties in the form of evolving variance estimates. 
We evaluate our setup, 
\ifdefined\realdata \revisionm{both on synthetic and real data,} \fi to extrapolate motion of rolling ball(s) on bowls of varying shape and orientation, and on arbitrary heightfields using only images as input.
We report significant improvements over existing image-based methods both in terms of accuracy of predictions and complexity of scenarios; and report competitive performance with approaches that, unlike us, assume access to internal physical states.

%% file: intro.tex
\section{Introduction}\label{s:intro}

Animals can make remarkably accurate and fast predictions of physical phenomena in order to perform activities such as navigate, prey, or burrow. However, the nature of the mental models used to perform such predictions remains unclear and is still actively researched~\cite{mentalModel}.

In contrast, science has developed an excellent \emph{formal understanding} of physics; for example, mechanics is nearly perfectly described by Newtonian physics. However, while the constituent laws are simple and accurate, applying them to the description of a physical scenario is anything but trivial. First, the scenario needs to be \emph{abstracted} (\eg,  by segmenting the scene into rigid objects, deciding which equations to apply, and estimating physical parameters such as mass, linear and angular velocity, \etc). Then, prediction still requires the \emph{numerical integration} of complex systems of equations. It is unlikely that this is the process of mental modeling followed by natural intelligences.

In an effort to develop models of physics that are more suitable for artificial intelligence, in this work, we ask  whether a representation of the physical state of a mechanical system can be learned \emph{implicitly} by a neural network, and whether this can be used to perform accurate predictions efficiently (i.e., extrapolating to predict future events). To this end, we propose a new learnable representation with several important properties. First, the representation is not handcrafted, but rather \revision{\emph{automatically induced from visual observations} using supervision from known physical quantities such as positions and angular velocities}. Second, the representation is \emph{distributed} and can model physical interactions of objects with \emph{complex surrounding}, such as irregularly-shaped ground. Third, despite its distributed nature, the representation can model a number of \emph{interacting discrete objects} such as colliding balls, without the need of ad-hoc components such as collision detection subnetworks. Fourth, since physical predictions integrate errors over time and are thus inherently ambiguous, the representation produces \emph{robust probabilistic predictions} which model such ambiguity explicitly. Finally, through extensive evaluation,  we show that the representation performs well for both {\em extrapolation} and {\em interpolation} of mechanical phenomena.

Our paper is not the first that looks at learning to predict mechanical phenomena using deep networks. \revisionm{In particular, inducing a physical representation automatically from visual data and handling object interactions were explored in}~\citep{fragkiadaki2015learning,NIPS2017_7040}. \revisionm{However, in this paper we propose a method that combines these benefits, with the main technical novelty of a distributed tensor representation for the physical state. The latter enables us to consider more complex environments.} 

Earlier, the recent Neural Physics Engine (NPE) of~\cite{chang2016compositional} uses a neural network to learn the state transition function of mechanical systems. Differently from ours, their state is handcrafted and includes physical parameters such as positions, velocities, and masses of rigid bodies. While NPE works well, it still requires to abstract the physical system manually, by identifying the objects and their physical parameters, and by explicitly integrating such parameters. \revision{In practice, this requires an extensive knowledge of the environment that, in turn, can bring in more estimation errors} \citep{yu2016more}. In contrast,  our abstractions are entirely induced from external observations of object motions. Hence, our system implicitly discovers any hidden variable or state required to perform tasks such as \emph{long-term physical extrapolation} in an optimal manner. Furthermore, the integration of physical parameters over time is also implicit and performed by a recurrent neural network architecture. This is needed since the nature of the internal state is undetermined; it also has a major practical benefit as, as we show empirically, the system can be trained to not only extrapolate physical trajectories, but also to interpolate them. Remarkably, interpolation is still obtained by computing the trajectory in a feed-forward manner, from the first to the last time step, using the recurrent model.

Another significant difference with NPE is in the fact that our system uses \emph{visual observations} to perform its predictions. In this sense, the work closest to ours is the \emph{Visual Interaction Networks} (VIN) of~\cite{NIPS2017_7040}, which also use visual input for prediction. However, our system is significantly more advanced as it can model the interaction of objects with complex and irregular terrain. We show empirically that VIN is not very competitive in our more complex experimental setting.

There are also several aspects that we address for the first time in this paper. Empirically, we push our model by considering scenarios beyond the `flat' ones tackled by most recent papers, such as objects sliding and colliding on planes, and look for the first time at the case of ball(s) rolling on non-trivial 3D shapes (e.g., bowls of varying shape and orientation, or terrains modeled as arbitrary heightfields), where both linear and angular momenta are tightly coupled. 
\ifdefined\realdata
\revisionm{Our method is evaluated on both \emph{synthetic and real} data using the \textsc{Roll4Real} dataset of} \cite{EhrhardtEtAl:UnsupervisedIntuitivePhysics:ACCV:18}. 
\fi 
\revisionm{We also increased the complexity of the task by training models that simultaneously estimate \emph{positions and angular velocity}.}
Furthermore, since physical extrapolation is inherently ambiguous, we allow the model to explicitly estimate its prediction uncertainty by estimating the variance of a Gaussian observation model. We show that this modification further improves the quality of long-term predictions. \revision{While our work builds on previous research of}  \cite{LearningPhysicalPredictor:emmv:2017} and  \cite{LearningMechanics:emvm:2017},    \revision{in this paper we propose a much more complex set of experiments including the irregularly shaped heightfield and multiple balls experiment as well as stronger baselines that highlight the performances of our models. Furthermore, a more careful study of the various results presented has been conducted and exhaustively discussed}.

The rest of the paper is organized as follows. The relation of our work to the literature is discussed in~\cref{s:related}. The detailed structure of the proposed neural networks is given and motivated in~\cref{s:method}. These networks are  tested on a large dataset of simulated physical experiments described in~\cref{s:phys} and extensively evaluated and contrasted against related works in~\cref{s:exp}. We conclude by discussing current limitations and directions for future investigation in \cref{s:conclusions}.

%% file: relatedWork.tex
\section{Related Work}\label{s:related}

We address the problem of training deep neural networks that can perform long-term predictions of mechanical phenomena while learning the required physical laws implicitly, via empirical and visual observation of the motion of objects. This research is thus related to a number of recent works in various machine learning sub-areas, discussed next.

\paragraph{Learning intuitive physics}

\cite{battaglia2013simulation} are one of the first to consider `intuitive' physical reasoning; their aim is to answer simple qualitative questions related to rigid body processes, such as determining whether a certain tower of blocks is likely to fall or not. They approach the problem by using a sophisticated physics engine that incorporates all required knowledge about Newtonian physics \emph{a-priori}. More recently, \cite{Mottaghi_2016_CVPR} used static images and a graphics rendering engine (Blender) to predict motion and forces from a single RGB image. Motivated by the recent success of deep learning for image analysis (\eg, \cite{krizhevsky2012imagenet}), they trained a convolutional neural network to predict such quantities and used it to produce a ``most likely motion,'' rendering it using a traditional computer graphics pipeline. With a similar motivation, \cite{lerer2016learning} and~\cite{li2016visual} also applied deep networks to predict the stability of towers of blocks purely from images. These approaches demonstrated that such networks can not only predict instability, but also pinpoint the source of such instability, if any. Other approaches such as \cite{NIPS2016_6113} or \cite{denil2016learning} have attempted to learn intuitive physics of objects through manipulation; however, their models did not aim to capture the underlying dynamics of the systems.

\paragraph{Learning physics} 

The work by~\cite{Galileo:NIPS:2015} and its extension~\cite{phys101} propose methods to learn physical properties of scenes and objects. \cite{Galileo:NIPS:2015} use an MCMC-sampling based approach that assumes complete knowledge of the physical equations necessary to estimate physical parameters. In~\cite{phys101}, a deep learning based approach was used instead of MCMC, albeit still explicitly encoding physics in a simulator. Physical laws were also explicitly incorporated in the model by~\cite{Stewart2016LabelFreeSO} to predict the movement of a pillow from unlabelled data. Their method was, however, only applied to a fixed small number of future frames. Work of \cite{yu2016more} \revision{proposing a high fidelity dataset for planar pushing reveals that what might be thought of as a simple task, to push an unknown  object  to  a  desired  position,  remains a challenging task in robotics, and is generally better explained by stochastic models than by estimating and modelling the physical world.}

The research performed by~\cite{battaglia2016interaction} and~\cite{chang2016compositional} focused on dynamics and attempted to partially substitute the physics engine with a neural network that captures a selection of relevant physical laws. Both approaches were able to use such networks to accurately predict updates for the physical state of the world. Although results are plausible and promising, \cite{chang2016compositional} suggest that long-term predictions remain difficult. Furthermore, in both approaches, their neural networks only predict instantaneous updates of physical parameters that are then explicitly integrated. In contrast, in this work propagation is implicit and applies a recurrent neural network architecture to an implicit representation of the world.

Closer to our approach, \cite{fragkiadaki2015learning} and \mbox{\cite{NIPS2017_7040}} attempted to learn an internal representation of the physical world from images. In addition to observing images, it is also possible to \emph{generate} them as \cite{fragkiadaki2015learning} learn to perform long-term extrapolation more successfully. The work of \cite{fragkiadaki2015learning} \revisionm{particularly differs from ours in this last point and the nature of its internal representation (vector representation in their work \emph{vs} tensor in ours) which, as demonstrated in} \ref{sec:res}, \revisionm{is essential for our method}. Similarly \cite{wu17learning} \revision{also used a physics engine and a renderer to make future predictions. In both cases, image generation can be seen as a constraint that avoids the over time degeneration of the internal representation of dynamics. However, these approaches need exhaustive and exact knowledge of every object and the environment, information generally not accessible in real life scenarios}.  The work of~\cite{NIPS2017_7040} extends the Interaction Network by \cite{battaglia2016interaction} to propagate an implicit representation of the dynamics of objects, obtaining a Visual Interaction Network (VIN). While their approach is the closest to ours, it has various limitations including not modeling the interaction with complex environments and the relatively small size of the input images. The Predictron by \cite{DBLP:journals/corr/SilverHHSGHDRRB16} also propagates a tensor state, but suffers from the same drawbacks. \cite{LearningMechanics:emvm:2017} showed, how long-term extrapolation models can be trained for one object moving on smooth analytic surfaces, such as ellipsoids.

\paragraph{Approximating physics for plausible simulation}

Several authors focused on learning to perform plausible physical predictions, for example to generate realistic future frames in a video~\cite{CNNFluid2016,jeong2015data}, or to infer rigid body collision parameters from monocular videos~\cite{MonszpartEtAl:SMASH:2016}. In these approaches, physics-based losses are used to learn plausible yet not necessarily accurate results, which may be appropriate for tasks such as rendering and animation.
\cite{battaglia2016interaction} also use a loss that captures the concept of energy conservation. The latter can be seen as a way to incorporate knowledge about physics \emph{a-priori} into the network, which differs from our goal of learning any required physical knowledge from empirical observations.

\paragraph{Learning dynamics}

Physical extrapolation can be performed without integrating physical equations explicitly. For example, LSTMs~\cite{Hochreiter:1997:LSM:1246443.1246450} were used  to make accurate long-term predictions in human pose estimation~\cite{longterm} and in simulated environments \cite{oh2015action,recurrentenv}. Propagation can also be done using simpler convolutional operators; \cite{visualdynamics16}, in particular, used these to generate possible future frames given a single static image and~\cite{debrabandere16dynamic} applied it to the moving MNIST dataset for long-term prediction. The work by~\cite{OndruskaAAAI2016} and \cite{LearningPhysicalPredictor:emmv:2017} also showed that an internal representation of dynamics can be propagated through time using a simple deep recurrent architecture. \cite{NIPS2017_7246} \revision{demonstrated that information about dynamics can be used to efficiently cluster different observed shapes.}
Our work builds on their success, and propagates a tensor-based state representation instead of a vector-based one. Using spatial convolutional operators allows for knowledge to be stored and propagated locally \mbox{\textit{w. r. t.}} the object locations in the images.

%% file: method.tex
\section{Method}\label{s:method}

\begin{figure}[t]
    \centering
    \begin{overpic}[width=0.6\linewidth]{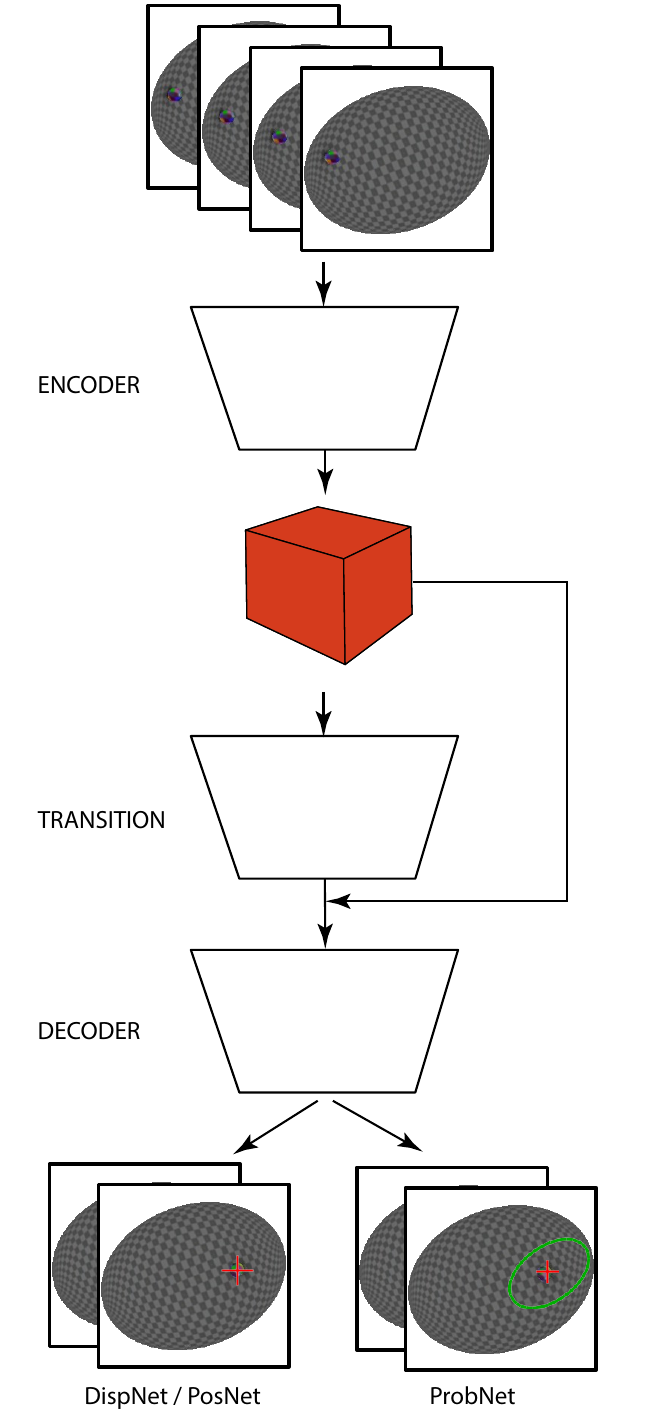}
    \put(5,101){\small Input images $t=0\ldots3$}
    \put(19,43){\large $\phi_\text{trans}$}
    \put(20,28){\large $\phi_{dec}$}
    \put(20,72.5){\large $\phi_{enc}$}
    \put(9,52){\scriptsize $3\times3\times N_f$}
    \end{overpic}
    \caption{\textbf{Overview of our proposed pipeline.} The first four images of a sequence first pass through a partially pre-trained feature encoder network to build the concept of physical state. It then recursively passes through a transition layer to produce long-term predictions about the future states of the objects. It is then decoded to produce state estimates. While our \TNet and \PTNet models are trained to regress the next states, the \PNet model trained with the log-likelihood loss is also able to handle the notion of uncertainty thanks to its extended state space. Note here that only one object is considered, extension for multiple objects is discussed in \cref{pare:ext}.}
    \label{fig:over}
\end{figure}
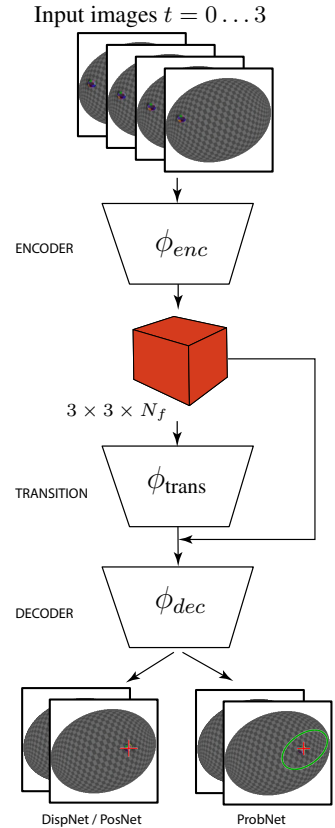

In this section, we propose a novel neural network model to make predictions about the evolution of a mechanical system from visual observations of its initial state. In particular, this network, summarized in~\cref{fig:over}, can predict the motion of one or more rolling objects accounting for variations in the 3D geometry of the environment.

Formally, let $y_t$ be a vector of physical quantities that we would like to predict at time $t$, such as the position of one or more objects. Physical systems satisfy a Markov condition, in the sense that there exists a state vector $h_t$ such that (i)~measurements $y_t=g(h_t)$ are a function of the state and (ii)~the state at the next time step $h_{t+1}=f(h_t)$ depends only on the current value of the state $h_t$. Uncertainty in the model can be encoded by means of observation $p(y_t|h_t)$ and transition $p(h_{t+1}|h_t)$ probabilities, resulting in a hidden Markov model.

State-only methods, such as the Neural Physics Engine (NPE) by~\cite{chang2016compositional} start from an handcrafted definition of the state $h_t$. For instance, in order to model a scenario with two balls colliding, one may choose $h_t$ to contain the position and velocity of each ball. In this case, the observation function $g$ may be as simple as extracting the position components from the state vector. It is then possible to use a neural network $\phi$ to approximate the transition function $f$. In particular, \cite{chang2016compositional} suggest that it is often easier for a network to predict a rate of change $\Delta_t = \phi(h_t)$ for some of the physical parameters (\eg, the balls' velocities), which can be used to update the state using a hand-crafted integrator $h_{t+1} = \tilde f(h_t, \Delta_t)$.

While this approach works well, there are several limitations. First, even if the transition function is learned, the state $h_t$ is defined by hand. Even in the simple case of the colliding balls, the choice of state is ambiguous; for example, one could include in the state not only the position and velocity of the balls, but also their radius, mass, elasticity, friction coefficients, \etc  Learning the state as well has the significant benefit of making such choices automatic. Second, training a transition function requires knowledge of the state values, which may be difficult to obtain except in the case of simulated data. Third, in order to use such a system to perform predictions, one must know the initial value of the state $h_0$ of the system, whereas in many applications one would like to start instead from \emph{sensory inputs} $x_t$ such as images~\cite{fragkiadaki2015learning}.

We propose here an approach to address these difficulties. We assume that the state $h_t$ is a \emph{hidden variable}, to be determined as part of the learning process. Since the $h_t$ cannot be observed, the transition function $h_{t+1}=f(h_t)$ cannot be learned directly as in the NPE. Instead, state and transitions must be inferred jointly as a \emph{good explanation} of the observed physical measurements $y_t$. Any integrator involved in the computation of the transition function is implicitly moved \emph{inside the network}, which is a recurrent neural network architecture. In our experiments (\cref{s:exp}), we show that the added flexibility of learning an internal state representation and its evolution automatically allows the system to scale well to the complexity of the physical scenario.

Since the evolution of the state $h_t$ cannot be learned by observing measurements $y_t$ in isolation, the system is supervised using sequences $\y_{[0,T)}=(y_0,\dots,y_{T-1})$ of observations. This is analogous to a Hidden Markov Model (HMM), which is often learned by maximizing the likelihood of the observation sequences after marginalizing the hidden state.%
\footnote{
Formally, a Markov model is given by $p(\y_{[0,T)},\h_{[0,T)}) = p(h_0) p(y_0|h_0) \prod_{t=0}^{T-2} p(h_{t+1}|h_{t})p(y_{t+1}|h_{t+1})$; traditionally, $p$ can be learned as the maximizer of the log-likelihood $\max_p E_y[\log E_\h[p(\y,\h)]]$, where we dropped the subscripts for compactness. Learning to interpolate/extrapolate can be done by considering subsets $\bar \y \subset \y$ of the measurements as given and optimizing the likelihood of the conditional probability $\max_p E_\y[\log E_\h[p(\y,\h|\bar \y)]]$.}
As an alternative learning formulation, we propose instead to consider the problem of long-term predictions starting from an initial set of observations. Not only this is more directly related to applications, but it has the important benefit that predictions can be performed equally well from initial observations of the physical quantities $y_t$ \emph{or} of some other sensor reading $x_t$, such as images.

Our system is thus based on learning three modules: 
(i)~an {\em encoder} function that estimates the state $h_t=\phi_\text{enc}(\x_{(t-T_0,t]})$ from the $T_0$ most recent sensor readings (alternatively $h_t=\phi_\text{enc}(\y_{(t-T_0,t]})$ can use the $T_0$ most recent physical observations); 
(ii)~a {\em transition} function $h_{t+1}=\phi_\text{trans}(h_{t})$ that evolves the state through time; and 
(iii)~a {\em decoder} function that maps the state $h_t$ to a physical observation $y_t=\phi_\text{dec}(h_t)$, and in some case an uncertainty associated.  The rest of the section discusses the three modules, encoder, transition, and decoder maps, as well as the loss function used for training. Further technical details can be found in \cref{s:exp}.

\subsection{Encoder Map: from images to state}

The goal of the encoder map is to take $T_0$ consecutive video frames observing the initial part of the object motion and to produce an estimate $h_0 = \phi_\text{enc}(\bx_{(-T_0,0]})$ of the initial state of the physical system. In order to build this encoder, we follow~\mbox{\cite{fragkiadaki2015learning}} and concatenate the RGB channels of the $T_0$ images in a single $H_i \times W_i \times 3T_0$ tensor. The latter is passed to a convolutional neural network $\phi_\text{enc}$ outputting a feature tensor $s_0\in \mathbb{R}^{H\times W\times C}$, used as internal representation of the system's state. Note that this representation is spatially distributed and differs from the concentrated vector representation of the VIN of~\cite{NIPS2017_7040}. In the experiments, we will show the advantage of using a tensorial representation in modeling complex environments. We also augment our tensor representation with a state vector $p_t\in\mathbb{R}^n$, so that the state is the pair $h_t=(s_t,p_t)$. In deterministic cases, $n=2$ and $p_t$ is the 2D projection of the object's location on the image plane. For multiple objects (see \cref{pare:ext}) this state is computed for each object independently.

\begin{table}[b!]
\center
\caption{\textbf{Neural network variants.}}\label{t:variants}
\vspace{1em}
\small
\begin{tabular}{|C{4em}|c|c|c|}
\hline
Name    & $p_t$ regression & $p_{t+1}$ & output and loss \\
\hline
\TNet   & incremental & $ p_t + \phi_p(s_t)$ & deterministic \\
\PNet   & incremental & $p_t + \phi_p(s_t)$ & probabilistic \\
\PTNet  & direct  & $\phi_p(s_t)$ & deterministic \\
\hline
\end{tabular}
\end{table}

\subsection{Transition Map: evolving the state}

The state $h_t$ is evolved through time by learning the transition function $\phi_\text{trans} : h_t \mapsto h_{t+1}$. Since the initial state $h_0$ is obtained from the encoder map, the state at time $t$ can be written as,  $h_t = \phi^t_\text{trans}(\phi_\text{enc}(\bx_{(-T_0,0]}))$. 

More in detail, the distributed state component $s_t$ is updated by using a convolutional network $s_{t+1} = \phi_s(s_t)$. The concentrated component $p_t$ is updated incrementally as $p_{t+1} = p_t + \phi_p(s_t)$, where $\phi_p(s_t)$ is estimated using a single layer perceptron regressor from the distributed representation. Combined, the state update can be written as,  
\vspace*{-.1in}
$$(s_{t+1},p_{t+1}) = \phi_\text{trans}(s_t,p_t) = (\phi_s(s_t),p_t + \phi_p(s_t)).$$
Inspired by the work of \cite{NIPS2017_7040}, we also consider an alternative architecture where $p_t$ is estimated directly from $s_t$ rather than incrementally. In order to do so, the location $x$ and $y$ of each pixel is appended as feature channels $C+1$ and $C+2$ of the distributed state tensor $s_t$, obtaining an augmented tensor $\operatorname{aug}_{xy}(s_t)$. Then the object's position $p_t$ is estimated by a two-layer perceptron $p_t = \phi_p(\operatorname{aug}_{xy}(s_t))$.

\subsection{Decoder Map: from state to probabilistic predictions}\label{subsec:decoder}

For deterministic models, the projected object position $p_t$ is part of the neural network state, the decoder map $\hat y_t = \phi_\text{dec}(s_t,p_t) = p_t$ simply extracts and returns that part of the state. Training optimizes the average $L^2$ distance between ground truth $y_t$ and predicted $\hat y_t$ positions $\frac{1}{T} \sum_{t=0}^{T-1} \|\hat y_t - y_t\|^2$.

In addition to this simple scheme, we also consider a more robust variant based on probabilistic predictions. In fact, the extrapolation error accumulates and increases over time, and the $L^2$-based loss may be dominated by outliers, unbalancing learning. Hence, we modify the model to explicitly and dynamically express its own prediction uncertainty by outputting the mean and variance $(\mu_t,\Sigma_t)$ of a bivariate Gaussian observation model. The $L^2$ loss is thus replaced with the negative log likelihood $- \frac{1}{T} \sum_{t=0}^{T-1} \log \mathcal{N}(y_t;\mu_t,\Sigma_t)$ under this model.

In order to estimate the Gaussian parameters $\mu_t$ and $\Sigma_t$, we extend the state component $p_t=(\mu_t,\lambda_{1,t}, \lambda_{2,t}, \theta_t)$ to include both the mean as well as the eigenvalues and rotation of the covariance matrix
$
\Sigma_t 
= 
R(\theta_t)^\intercal 
\operatorname{diag}(\lambda_{1,t},\lambda_{2,t})
R(\theta_t)
$.
In order to ensure numerical stability, eigenvalues are constrained to be in the range $[0.01,100]$ by setting them as the output of a scaled and translated sigmoid $\lambda_{i,t} = \sigma_{\lambda,\alpha}(\beta_{i,t})$, where $\sigma_{\lambda,\alpha}(z) = \lambda/(1 + \exp(-z)) + \alpha$. In the following, we will refer to this method as \PNet, whereas the other method estimated displacement without uncertainty will be referred to as \TNet. \cref{t:variants} summarize the different methods and their specificity.

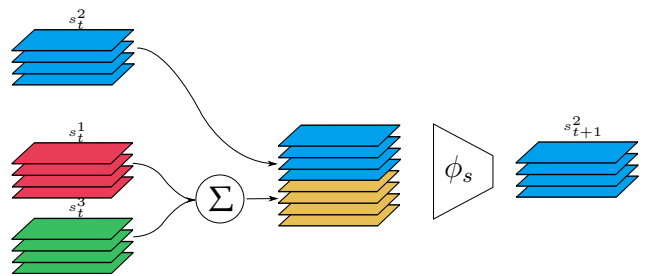
\begin{figure}[b!]
    \centering
    \begin{overpic}[width=\linewidth]{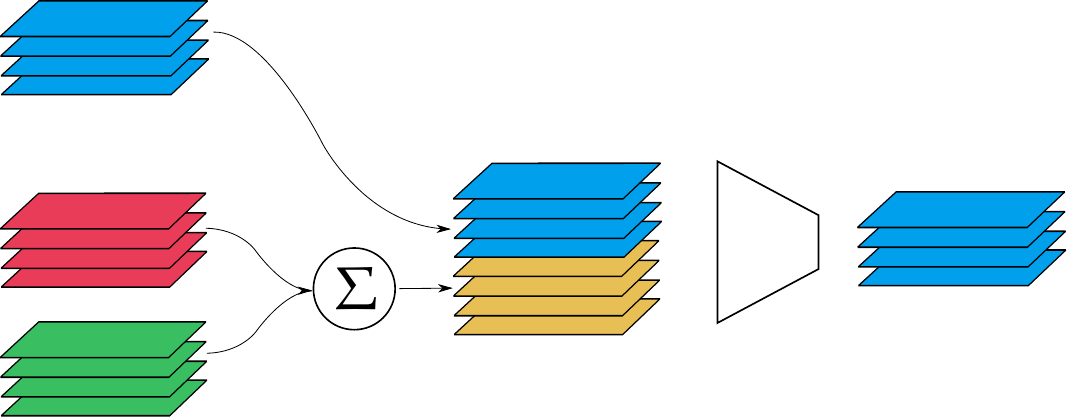}
    \put(9,40){\tiny $s_{t}^2$}
    \put(9,22){\tiny $s_{t}^1$}
    \put(9,10){\tiny $s_{t}^3$}
    \put(88,23){\tiny $s_{t+1}^2$}
    \put(69,15.5){\large $\phi_s$}
    \end{overpic}
    \vspace{0.3em}
    \caption{\textbf{Multiple object extension.} For each object (here object 2) we concatenate the state of this object with the addition of the other objects features. We then give this tensor to the module $\phi_s$ to obtain our new state $s_{t+1}^2$}
    \label{fig:multiple}
\end{figure}

\subsection{Extension to multiple objects}\label{pare:ext}

We now consider how the model described above can be extended to handle multiple interacting objects. This is more challenging as it requires to handle complex object interactions such as collisions.

In order to do so, for each object $o_i,$ $i=1,\dots,N_\text{objects}$ we consider a separate copy of the distributed state tensor $s^{o_i}_t$ (hence the overall state is $s_t=(s_t^{o_1},\dots,s_t^{o_{N_\text{objects}}})$). The encoder network $\phi_\text{enc}$ is thus modified to output a $H\times W\times {N_\text{objects}} C$ tensor. It is then split along the third dimension to produce $H\times W\times C$ tensor for each of the $N_{objects}$. We order objects \textit{w.\ r.\ t.} their color so that each feature is always responsible for the same object identified by its color. We recall here that this extension studies the ability of handling collisions of our model without any explicit module. We aim in the future to build more object agnostic representation.

The input of the transition module is also modified to take into account the interaction between objects. Focusing on an object $o_f$ with state $s^{o_f}_t$, the update is written as
$$
s^{o_f}_{t+1} = \phi_s
\left(s^{o_f}_t,\sum_{i\neq f}s^{o_i}_t\right)
$$
where the second argument is the sum of the state subtensors for all \emph{other} objects. Since the function $\phi_s$ is the same for all objects $o_f$, this ensures that object interactions are symmetric and commutative.
Note that, as opposed to methods such as~\cite{chang2016compositional}, no explicit collision detection module is implemented here. Instead, handling collisions is left to the discretion of the network. 

With this modification, the transition subnetwork is illustrated in~\cref{fig:multiple}. The rest of the pipeline is essentially the same as before and is applied independently to each object. The same network parameters are used for each application of a module regardless of the specific object.

%% file: experimentalSetup.tex
\section{Experimental Setup}\label{s:phys}

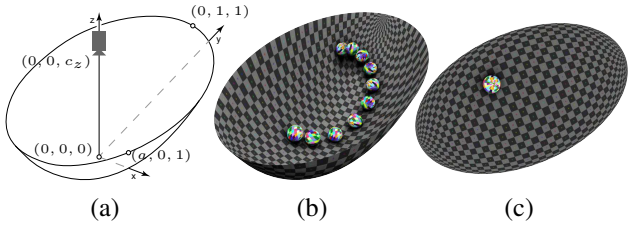
\begin{figure}[h!]
\centering
\begin{overpic}[width=\linewidth]{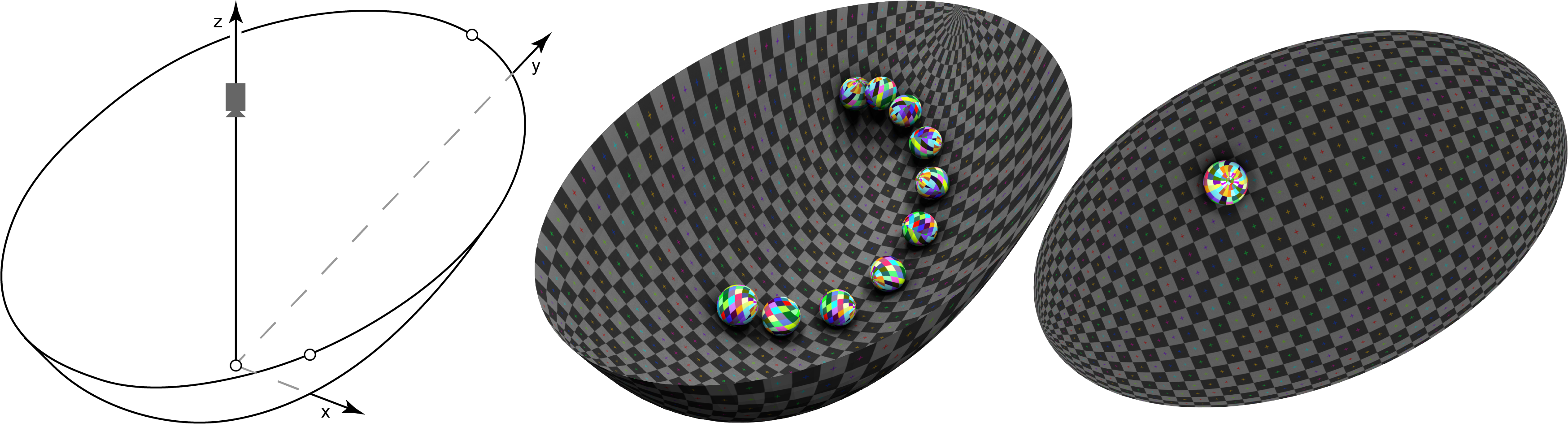}
\put(30,27){\tiny $(0,1,1)$}
\put(3,19){\tiny $(0,0,c_z)$}
\put(5,5){\tiny $(0,0,0)$}
\put(20.5,4){\tiny $(a,0,1)$}
\end{overpic}
\mbox{}\hfill(a)\hfill\hfill(b)\hfill\hfill(c)\hfill\mbox{}
\vspace{0.5em}
\caption{\textbf{Problem setup.} We consider the problem of understanding and extrapolating mechanical phenomena with recurrent deep networks. (a) Experimental setup: an orthographic camera looks at a ball rolling in a 3D bowl. (b) Example of a 3D trajectory in the 3D bowl simulated using Blender 2.78's OpenGL renderer. (c) An example of a rendered frame in the \bowltwoq experiment that is fed to our model as input. }
\label{fig:simulation_setup}
\end{figure}

Experiments were conducted on both real and synthetic datasets. In the synthetic experiments (\cref{fig:simulation_setup}), we consider two physical scenarios: spheres rolling on a 3D surface, which can be either a semi-ellipsoid with random parameters or a continuous randomized heightfield. When the semi-ellipsoid is isotropic (\ie\ a hemisphere) we refer to it as \bowloneq, and in the more general case as \bowltwoq  (see~\cref{t:one}), whereas the heightfield scenario is referred to as `Heightfield.'

\subsection{\bowlone and \bowltwo scenarios}

The symbol $\mathbf{p} = (p_x,p_y,p_z)\in\mathbb{R}^3$ denotes a point in 3D space or a vector (direction). The camera center is placed at location $(0,0,c_z)$, $c_z>0$ and looks downward along vector $(0,0,-1)$ using orthographic projection, such that the point $(p_x,p_y,p_z)$ projects to pixel $(p_x,p_y)$ in the image.

Thus, the \bowltwoq is the bottom half of an ellipsoid of equation $x^2/a^2 + y^2 + (z-1)^2=1$ with its axes aligned to the xyz axes and its lowest point corresponding to the origin. For the \bowltwoq scenario, the ellipsoid shape is further varied by sampling $a \in U[0.5,1]$ for the  ($a=1$ for the \bowloneq scenario) and by rotating the resulting shape randomly around the $z$-axis. Both \bowloneq and \bowltwoq are rendered by mapping a checker board pattern to their 3D surface (to make it visible to the network). 

The rolling object is a ball of radius $\rho \in \{0.04, 0.225\}$. The ball's center of mass at time $t$ is denoted as $\mathbf{q}^t = (q_x^t,q_y^t,q_z^t)$, which, due to the orthographic projection, is imaged at pixel $(q_x^t,q_y^t)$. 
The ball has a fixed multi-color texture attached to its surface, so it appears as a painted object. The texture is used to observe the object rotation. We study the impact of being able to visually observe rotation by re-rendering the single ball experiments with  a uniform white color (see \mbox{Table~\ref{t:no_text}}). In the multi-object experiments, instead, each ball has a constant, distinctive diffuse color (intensity 0.8) with Phong specular component (intensity 0.5). 
We initially position the ball at angles $(\theta,\phi)$ with respect to the the bowl center, where the elevation $\theta$ is uniformly sampled in the range $\theta \in U[-9\pi/10, -\pi/2]$ and the azimuth $\phi \in U[-\pi,~\pi]$. The minimum elevation is set to $-9\pi/10$ to avoid starting the ball at the bottom of the bowl. Due to friction, at the end of each experiment the ball rests at the bottom of the bowl.

The initial orientation of the ball (relevant for the multi-colored texture) is obtained by uniformly sampling its xyz Euler angles in $[-\pi,\pi]$. The ball's initial velocity $\bv$ is obtained by first sampling $v_x,v_y$ uniformly in the range $U[5,10]$, assigning each of $v_x,v_y$ a random sign ($\sim 2\mathcal{B}\left(0.5\right)-1$), and then by projecting the vector $(v_x,v_y,0)$ to be \emph{tangential} to the bowl's surface. In the multi-object \bowltwoq scenario, in order to achieve more interesting motion patterns, the magnitude of the initial velocities is set uniformly in the range $U[10, 15]$; if, after simulation, a ball leaves the bowl due to a collision or excessive initial velocity, the scene is discarded. Sequences are recorded until all objects stop moving. Short sequences (less than 250 frames) are discarded as well. The average angular velocity computed over all `Bowl' scenes was $5.94$ radian/s.

\begin{figure}[ht!]
    \centering
    \includegraphics[width=\columnwidth]{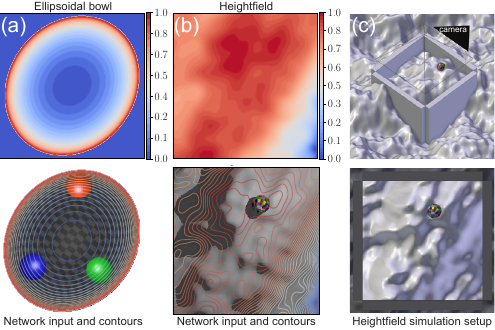}
    \vspace{0.2em}
    \caption{\textbf{Experimental setups.} (a) \bowltwoq experiment setup, depth map on the top, network input with isocontours at the bottom. We create the dataset by varying the ellipsoid's main axis ratio and orientation, and the starting position and velocity of the balls. (b-c) `Heightfield' rendering setup. Each sequence is generated using a random translation and rotation of the fixed heightfield geometry. Walls ensure the automatically generated sequences are long enough. A randomly positioned area light presents additional generalization challenges to the network. }
    \label{f:ellipse_contours}
\end{figure}

Note that, while some physical parameters of the ball's state are included in the \emph{observation} vector $\y^\alpha_{[-T_0,T)}$, these are \emph{not} part of the state $h$ of the neural network, which is inferred automatically. The network itself is tasked with \emph{predicting} part of these measurements, but their meaning is not hardcoded.

\subsubsection*{Simulation details}

For efficiency, we extract multiple sub-sequences $\bx^\alpha_{[-T_0,T)}$ form a single longer simulation (training, test, and validation sets are however completely independent). The simulator runs at 120fps for accuracy, but the data is subsampled to 40fps. We use Blender 2.78's OpenGL renderer and the Blender Game Engine (relying on Bullet 2 as physics engine). The ball is a textured sphere with unit mass.
The simulation parameters were set as: \mbox{max physics steps = 5}, \mbox{physics substeps = 3}, \mbox{max logic steps = 5}, \mbox{fps = 120}. Rendering used white environment lighting (energy = 0.7) and no other light source in the \bowloneq case, environment energy = 0.2, and a spotlight at the location of the camera in the \bowltwoq case. 
We used 70\% the data for training, 15\% for validation, and
15\% for test, 12500 sequences in the \bowloneq/\bowltwoq experiments and 6400 in the `Heightfield' case.
During training, we start observation at a random time while it is fixed for test. The output images were stored as $256\times256$ color JPEG files.
For multiple objects in the ellipsoid experiment, we set the elasticity parameter of the balls to 0.7 in order to get a couple of collisions before they settle in the middle of the scene.

\subsection{Heightfield scenario}

An important part of our experiments involve randomly generated continuous heightfields. Long-term motion prediction on random heightfields represent a tougher challenge, since solely observing the motion of the object at the beginning of the sequence does not contain enough information for successful mechanical predictions. In contrast to the \bowltwoq cases, where the 2D shape that the container occupies in the image is theoretically enough to infer the analytical shape of the local surface at any future 3D point of interest, in the `Heightfield' case the illumination conditions of the surface have to be parsed. Furthermore, a more elaborate understanding about the interaction between surface and 3D rolling motion has to be developed. 

Similar to the ellipsoid cases, we generate randomized sequences of a ball rolling on a random (heightfield) surface. We approximate random heightfields by generating a large (8 $\times$ 8) Improved Perlin noise texture and applying it as a displacement map to a highly tessellated plane. For each scene, we uniform randomly rotate and translate the plane so that a different part (2.5 $\times$ 2.5) of the heightfield is visible under the static camera. In order to generate motion sequences of enough length for long-term extrapolation, we also surround the camera frustum with perfectly elastic walls (see \mbox{Fig.~\ref{f:ellipse_contours}c}).
The noise texture has a scale parameter, which we vary between 0.7 (fairly planar) and 0.2 resulting in high curvature surfaces that have holes comparable with the ball diameter. 
We set the surface elasticity to 0 in order to encourage the balls to roll and not bounce. The initial placement of the ball, similarly to the bowl case, is drawn from a 2D uniform distribution. Then, we use sphere tracing to push the ball onto the surface from the camera plane. We add a small random initial velocity ($U[2, 4$]), and similarly to the \bowloneq case, we project the initial velocity onto the local surface normal. The average angular velocity computed over all `Heightfield' scenes was $2.8$ radian/s.
The surface is lit with a small ($0.1\times0.1$) area light from a random location. We draw the 2D position of the light as %
$x, y \sim \left(2\mathcal{B}\left(0.5\right) - 1\right) \left(U\left[1, 1.5\right] \times U\left[1, 1.5\right]\right)$, with a fixed camera height $z=2$.

\ifdefined\realdata
\subsection{Real data}
\revisionm{Additonally, we experimented on real data. We evaluated our methods on the \textsc{Roll4real} dataset by} \cite{EhrhardtEtAl:UnsupervisedIntuitivePhysics:ACCV:18}.
\revisionm{The dataset consists of 1118 short 256 $\times$ 256 videos containing one or two balls rolling on three types of terrains: a flat pool table \textsc{PoolR}, a large ellipsoidal `bowl' \textsc{BowlR}, and an irregular height-field \textsc{HeightR}. More specifically, there are 151 videos (avg. 99 frames/video) for the \textsc{PoolR} dataset with one ball; 216 videos (522 frames/video).
For the \textsc{BowlR} dataset with one ball; 543 videos (avg. 356 frames/video) for the \textsc{HeightR} dataset with one ball; and 208 videos (avg. 206 frames/video) for the \textsc{HeightR} dataset with two balls.
More details about the dataset and the way to obtain ground-truth annotations can be found in} \cite{EhrhardtEtAl:UnsupervisedIntuitivePhysics:ACCV:18}.
\fi

%% file: results.tex
\section{Results and Discussions}\label{s:exp}

\begin{figure*}[th!]%
    \centering
    \includegraphics[width=\linewidth]{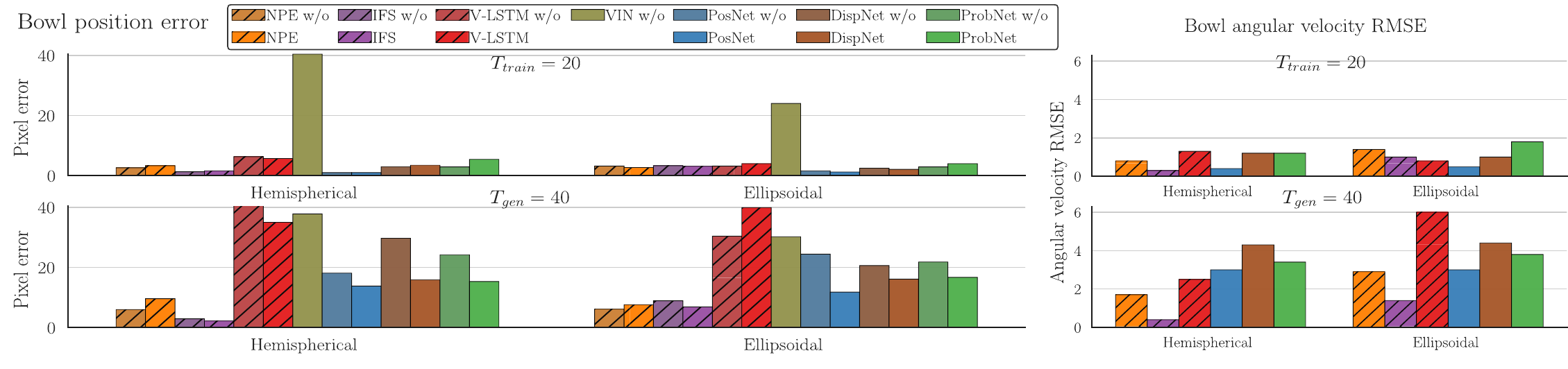}
    \caption{\textbf{Errors in bowls.} Pixel errors and angular velocity RMSE in radian/s (first two columns of \Cref{t:one}). Our method performs comparably to state based methods, which use ground truth state information for initialization compared to ours, which operates with \emph{visual} input. Hatched denotes non-visual input (\ie direct access to physical states).}
    \label{fig:bars_ellipse}
\end{figure*}%

\subsection{Baselines}

\paragraph{(i) Least squares fit} We compare the performance of our methods to two simple least squares baselines: Linear and Quadratic. In both cases, we fit least squares polynomials to the screen-space coordinates of the first $T=10$ frames, which are not computed but given as inputs. The polynomials are of first and second degree(s), respectively. Note, that being able to observe the first 10 frames is a large advantage compared to the networks, which only see the first $\ninputs=4$ frames.

\paragraph{(ii) NPE}  The NPE method and its variants were trained using available online code. We used the same training procedure as reported in \cite{chang2016compositional}. Additionally, we added angular velocities as input and regressed type of parameter. In the case of the \bowltwo, both scaling and bowl rotation angle are also given as input to the networks. In this case NPE's method carries forward the estimated states via the network. 
  
\paragraph{(iii) V-LSTM model} \revision{Inspired by the  models} of \protect\cite{fragkiadaki2015learning}, 
\revision{we developed an LSTM architecture as a baseline. The architecture is similar to the one in} \protect\cite{fragkiadaki2015learning},  \revisionm{as it reuses the exact same truncated pre-trained AlexNet encoder, and the same LSTM and decoder architecture with the following \textbf{two differences}:}
\revision{First, we do not regenerate images to produce a new input for the LSTMs, we rather used the last output of the LSTM. Second, the decoder only produces the next state estimate and not the 20 next ones to make it a fair comparison to our models. As the multiple ball version in the original paper required centering the frame of each independent object and regenerating images at every time step, we considered a simple reimplementation with one V-LSTM network for each of the individual objects. Thus we can take into account the entire frame without any need to regenerate images.}
            
\paragraph{(iv) VIN and IN From State (IFS)} Finally, we used VIN network and its state variant IN From State from \cite{NIPS2017_7040}. IFS is essentially a version of VIN where the propagation mechanism is the same but the first state vector is not deduced from visual observation but given as ground truth position and velocity as in the NPE. The VIN network uses downscaled $32\times32$ images. Both networks use training procedures as reported in \cite{NIPS2017_7040} with the exceptions that for IFS the learning rate was updated using our method (see \cref{sec:res}) and we rely on the first 4 states and 16 rolled out steps. As with NPE, angular velocity was also added to IFS input and regressed parameters. Scaling and rotation angle of the bowl were also given as input to the network in \bowltwo experiment. Note that VIN and our models work with images as direct observation of the world rather than perfect states, which represents a much more difficult problem whilst yielding a more general applicability. %
Physical properties are then deduced from the observations and integrated through our Markov model. Thus, these methods do not need a simulator to estimate parameters of the physical worlds (such as scaling and rotation angle) and can be trained on changing environments without requiring additional external measurements of the underlying 3D spaces.

\subsection{Results}\label{sec:res}

\paragraph{Implementation details} The encoder network $\phi_\text{enc}$ is obtained by taking the ImageNet-pretrained VGG16 network of \cite{Simonyan15} and retaining the layers up to \texttt{conv5} (for an input image of size $(H_i,W_i)=(128,128,3)$ this results in a $(8,8,N_f=512)$ state tensor $s_t$). In the 3 balls experiments, we replaced the last \texttt{conv5} layer with a convolutional layer of output $256\times3$ channels. Object features are thus obtained by splitting this last tensor along the channel dimension into $(8,8,N_f = 256)$ state tensor per object. The filter weights of all layers except \texttt{conv1} are retained for fine-tuning on our problem. The  \texttt{conv1} is reinitialized as filters must operate on images with $3T_0$ channels. The transition network $\phi_s(s_t)$ uses a simple chain of two convolution layers%
\footnote{\revisionm{We did not see the need to use an architecture incorporating a gating mechanism, such as a Conv-LSTM}~\cite{xingjian2015convolutional}, \revisionm{because in our case the transition function $\phi_s(s_t)$ does not observe new evidence after the first $T_0$ frames rendering the use of gating less useful}.}%
with $256$ and $N_f$ filters respectively, of size $3\times3$, stride 1, and padding 1 interleaved by a ReLU layer.Network weights are initialized by sampling from a Gaussian distribution. Additionally, angular velocity is always regressed from the state $s_t$ using a single layer perceptron.

Training uses a batch size of 50 using the first $T_\text{train}$ positions and angular velocity (or only position when explicitly mentioned) of each video sequence using RMSProp by \cite{Tieleman2012}. We start with a learning rate of $10^{-4}$ and decrease it by a factor of 10 when no improvements of the loss have been found after 100 consecutive epochs. Training is halted when the loss has not decreased after 200 successive epochs; 2,000 epochs were found to be usually sufficient for convergence. In every case the loss is the sum of the $L^2$ angular velocity loss and either $L^2$ position errors (\PTNet, \TNet) or likelihood loss (\PNet) (see \cref{subsec:decoder}). We omit the angular loss, when angular velocity is not regressed (labelled as ``\textit{* w/o ang.\ vel.}'' in the tables).

Since during the initial phases of training the network is very uncertain, the model using the Gaussian log-likelihood loss was found to get stuck on solutions with very high variance $\Sigma(t)$. To address this, we added a regularizer $\lambda \sum_t \det \Sigma(t)$  to the loss, with $\lambda=0.01$.

In all our experiments we used Tensorflow (\cite{tensorflow2015-whitepaper}) r1.3 on a single NVIDIA Titan X GPU.

\input{tabs/tab_oneball_synth.tex}

\subsubsection{Extrapolation}\label{ssec:experiments}

\paragraph{(i) Experiments using a single ball}
\Cref{t:one} compares the baseline predictors and the eight networks on the task of long term prediction of the object trajectory. All methods observed only the first $T_0=4$ inputs (either object states or simply image frames) except for the linear and quadratic baselines, and aimed to extrapolate the trajectory to $T_\text{gen}=40$ time steps. Predictions are ``long term'' relative to the number of inputs $T_0 \ll T_\text{gen}$. Note also that during training networks only observe sequences of up to $T_\text{train}\leq T_\text{gen}$ frames; hence, the challenge is not only to extrapolate physics, but to generalize \emph{beyond} extrapolations observed during training.

\paragraph{Quantitative evaluation} \Cref{t:one} reports the average errors at time $T_\text{train}=20$ and $T_\text{gen}=40$ for the different estimated parameters. Our methods outperform state-only approaches for predictions of up-to $T_\text{train}$ steps. For example, PosNet has a pixel error of 1.0/1.2/6.8 in the Hemispherical/Ellipsoidal/Heightfield scenarios vs 3.3/2.7/10.9 of NPE, 1.6/3.1/8.7 of IFS. This is non-trivial as our networks know nothing about physical laws a-priori, and observe the world through images rather than being given the initial ground-truth state values. On the other hand, our methods can, through images, better observe and hence model the underlying environments. The gap in the heightfield results, in particular, shows the value in observing the environments in this manner as we constantly out-perform state-only methods. Our methods also shown to make significantly better predictions compared to the other visual competitors. \revision{For instance, V-LSTM was unable to match the strong performance of our networks (pixel errors are 5.7/4.0/8.8 in the Hemispherical/Ellipsoidal/Heightfield scenarios respectively) highlighting the advantage of a spatially distributed tensor state representation as opposed to a vector one. As for the VIN network, it} failed to be able to model interactions between the object and its environment and performed poorly even on training regimes (40.4/24.0/42.6 respectively). 

\begin{figure}[t!]%
    \includegraphics[width=0.47\linewidth]{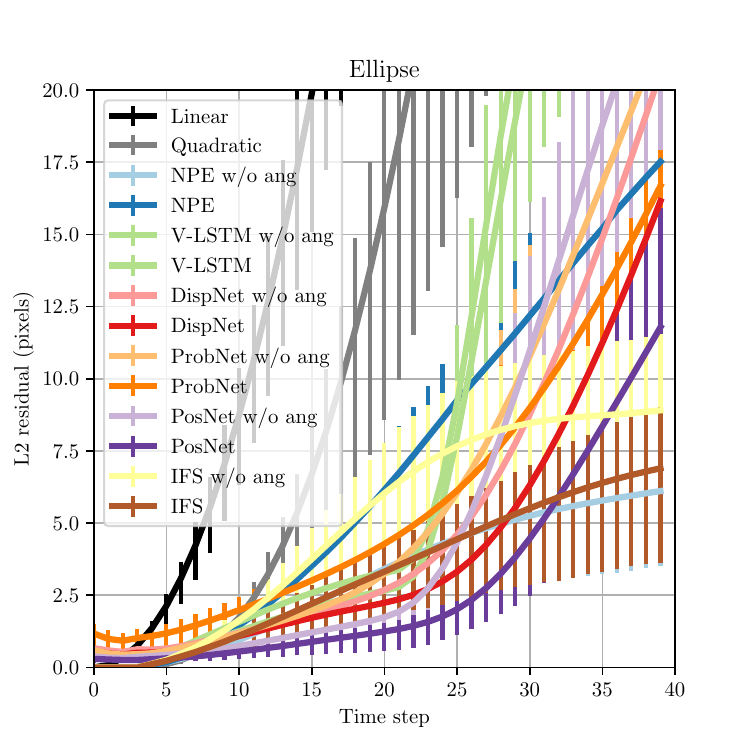}
    \includegraphics[width=0.47\linewidth]{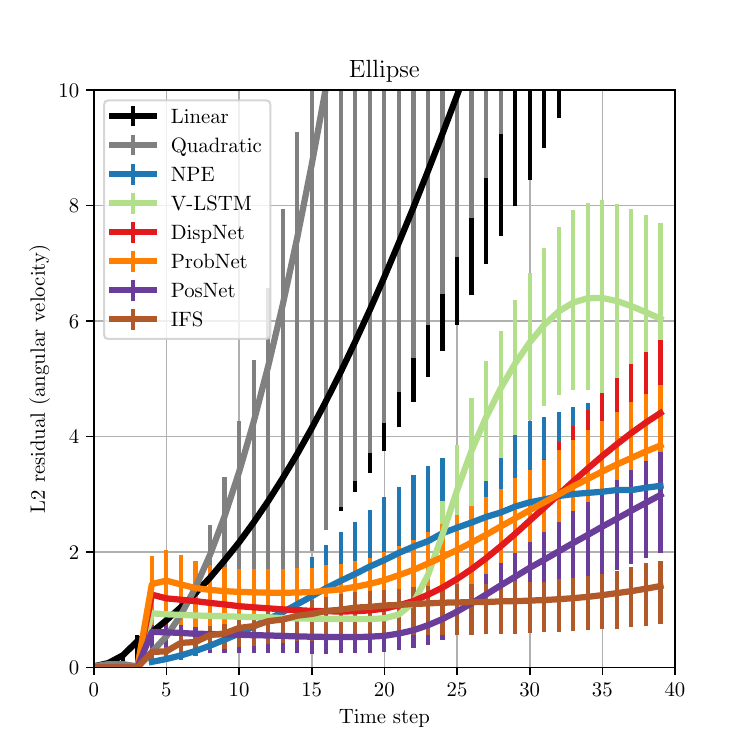}%
    \caption{\textbf{Errors evolution on Ellipsoidal Bowl} \revision{Position errors (left) and angular velocity error (right). We see that position and angular velocity errors degenerate outside training regimes (t=20) for all non state-only methods with an effect more tempered for our method. The impact is more moderate on angular velocity since its range is smaller than positions. Error bar shows 25th and 75th percentiles.}}%
    \label{fig:evolution}
\end{figure}

All methods can perform arbitrary long predictions. Our networks, which are only trained to predict the first $T_\text{train}$ positions, are still competitive with state-only methods (which only predict a transition function and hence implicitly generalize to arbitrarily lengths) even when predictions are generalized to $T_\text{gen}$ steps. In particular, while performances around $T_\text{gen}$ deteriorates, \PTNet provides very promising results, reaching nearly state-only models performances on the \bowltwoq experiments (11.8 pixel prediction error vs 6.1 of NPE). \cref{fig:evolution} \revision{shows the error evolution through time. This plot shows that in the long term our predictions seems to degenerate quicker than state-only methods outside training regimes but still remains more moderate than the other baselines.}

\begin{figure*}[t!]%
    \includegraphics[width=1.0\linewidth]{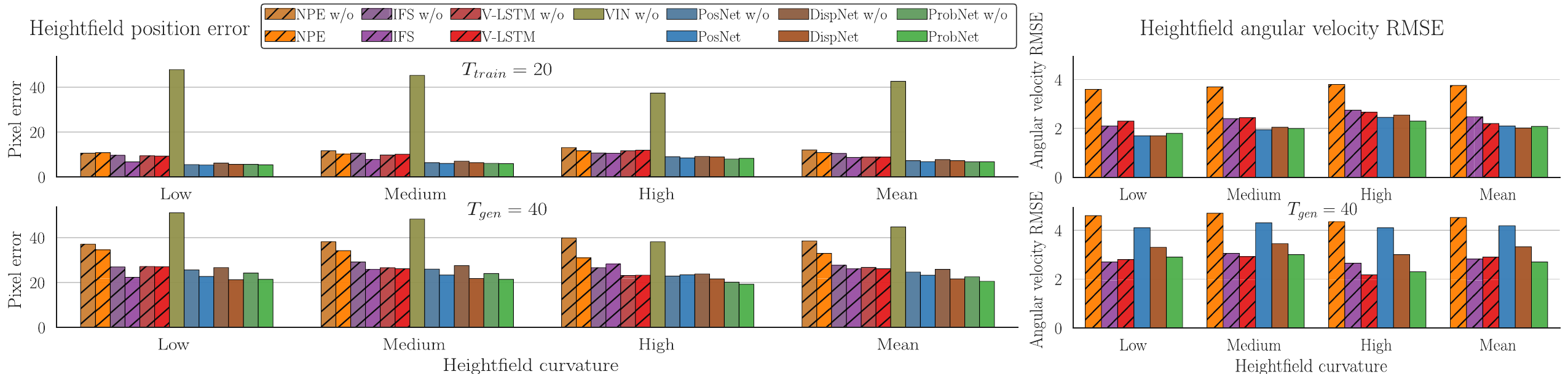}%
    \caption{\textbf{Errors on Heightfields.} Position errors (left) and angular velocity error (right) for trained (\Ttrain $=20$) and untrained (\Text $=40$) generalization on increasing difficulty heightfields ('Mean' is reported in the right column of \Cref{t:one}). Note, how angular velocity estimation helps position accuracy. Hatches denote non-visual methods.}%
    \label{fig:bars-heightfields}
\end{figure*}

We also note that learning to regress \emph{angular velocity} generally improve the ability of our models to predict \emph{position}, in particular when generalizing to $T_\text{gen}$ steps. For example, PosNet in the Ellipsoidal bowl reduces its position error from 24.4 to 11.8 at $T_\text{gen}$ when it is required to predict angular velocity during training. For further comparisons, see the similarly colored, adjacent bars in \mbox{\Cref{fig:bars_ellipse}~(left)} and \mbox{\Cref{fig:bars-heightfields}~(left)}). This is remarkable as angular velocity as such remains very challenging to predict. 

An interesting question is whether the model learns or not to measure angular velocity from images, or whether predicting this quantity during simply induces a better internal understanding of physics. To tease this effect out, we prevent the network from observing the ball spin by removing the texture on the ball. \Cref{t:no_text} shows that this results approximately in the same accuracy as the textured cases, indicating that angular velocity is not estimated visually. \revision{Our hypothesis is that angular velocity is estimated by exploiting the strong correlation between linear and angular velocities due to conservation of momentum}. 

Finally, introducing the probability-based loss in DispNet results in the ProbNet network. As shown in~\cref{t:one}, This change significantly outperforms the deterministic DispNet results in most cases.

\setlength{\tabcolsep}{0.4em}
\begin{table}[b!]
\caption{\textbf{Impact of ball texturing on prediction.} We compare the impact of ball texturing on predictions. Table layout and measures are same as \cref{t:one}. Results show that ball texture is rather ignored to make predictions.}
\vspace{0.7em}
\centering
\tiny 
\sisetup{detect-weight=true,detect-inline-weight=math,table-column-width=2.5em}
\begin{tabular}{|c|*4S[table-column-width=3.5em , table-format=-2.2]|*4S[table-column-width=3.5em ,table-format=-2.2]|}
\hline
    &   \multicolumn{4}{c|}{\bowltwo} & \multicolumn{4}{c|}{\bowltwo (no ball texture)}\\
    Method & 
    \multicolumn{4}{c|}{Errors (Perplexity)} &
    \multicolumn{4}{c|}{Errors (Perplexity)}\\
    &  \multicolumn{2}{c}{\Ttrain} & \multicolumn{2}{c|}{\Text} & \multicolumn{2}{c}{\Ttrain} & \multicolumn{2}{c|}{\Text}\\[0.1cm]
    & \multicolumn{1}{c}{\pos} 
    & \multicolumn{1}{c}{\av} 
    & \multicolumn{1}{c}{\pos} &\multicolumn{1}{c|}{\av}
    & \multicolumn{1}{c}{\pos} 
    & \multicolumn{1}{c}{\av} 
    & \multicolumn{1}{c}{\pos} &\multicolumn{1}{c|}{\av}\\
    \hline
    \PTNet \woa & 1.6 & \noentry & 24.4 & \noentry & 1.6 & \noentry & 23.7  & \noentry\\
    \hline 
    \PTNet & \bfseries 1.2 &  \bfseries 0.5 & \bfseries 11.8 & \bfseries 3.0 & \bfseries 1.1 & \bfseries 0.6 & \bfseries 12.7  & \bfseries 3.5\\
    \hline
    
    \TNet \woa & 2.5 & \noentry & 20.6 & \noentry & 1.7 & \noentry & 26.3  & \noentry \\
    \hline
    \TNet &  2.1 & 1.0 & 16.1 & 4.4 & 1.6 & 1.0 & 16.2 & 3.8\\
    \hline
    
    \PNet \woa &  2.9 & \noentry & 21.8 & \noentry   & 3.1 &\noentry & 24.0 &\noentry\\
               & {(32.1)} && {(54.0)} && {(5.0)} & & {(12.7)} &\\
    \hline
    \PNet & 4.0 & 1.8 & 16.7 &  3.8 & 4.3 & 1.3 & 15.0 & \bfseries 3.5\\
                   & {\bfseries (4.5)} && {\bfseries (9.3)} && {\bfseries (4.5)} && {\bfseries (8.2)} &\\
\hline  
\end{tabular}
\label{t:no_text}
\end{table}

\paragraph{(ii) Experiments using multiple balls}

\begin{figure*}
    \vspace{-2em}
    \centering
    \includegraphics[width=.875\textwidth]{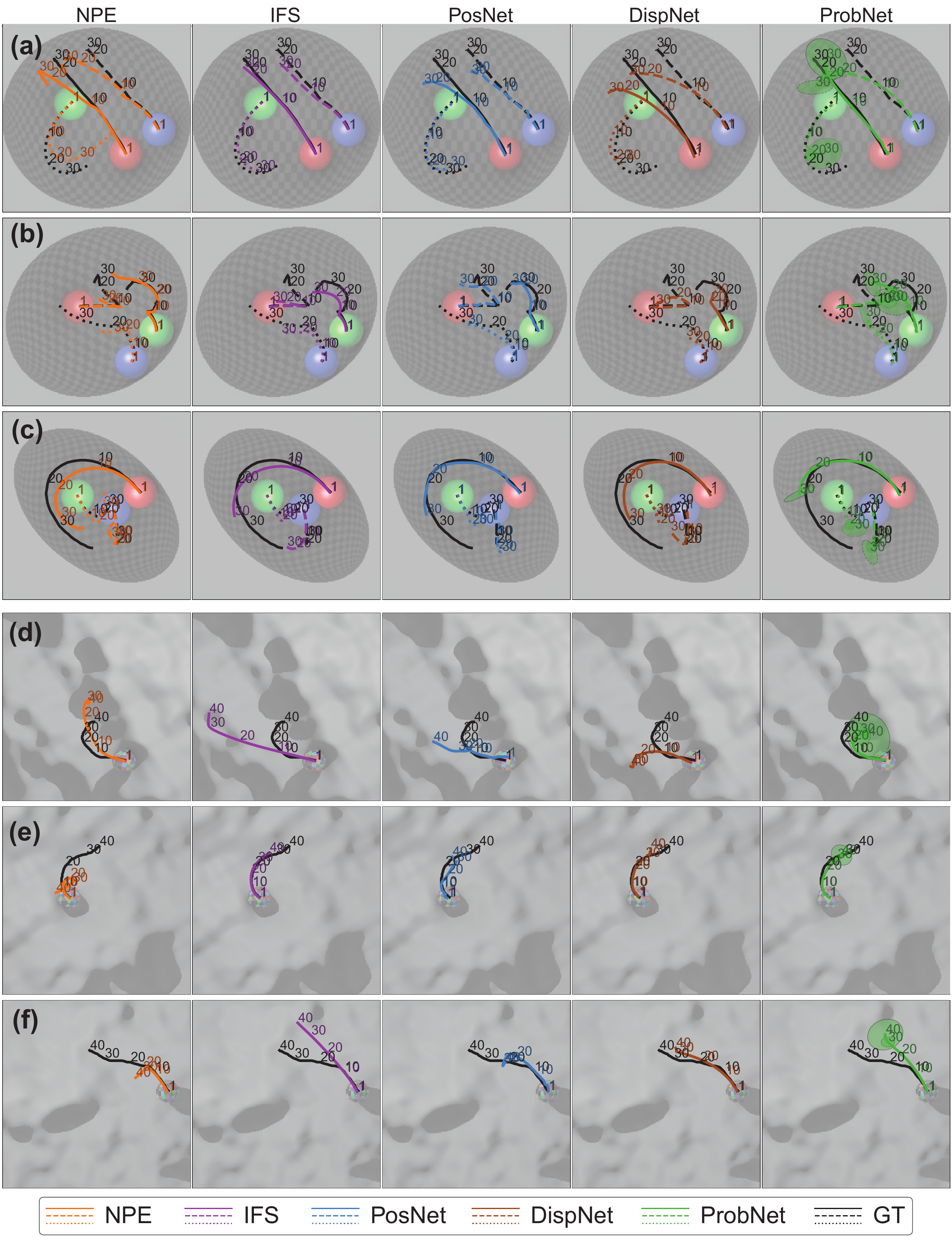}%
     \tiny{\caption{\textbf{\bowltwo and Heightfield extrapolations.} (a-c)~Example scene from the 3 balls in the \bowltwoq experiment. Extrapolation on multiple objects generalises well to 3 objects.
    Note how in (b) the collision of the red and green ball is predicted by our networks, solely by seeing the first 4 frames of the sequence. Remember, NPE and IFS start with the ground truth knowledge of the physical state of the objects.
    (d-f)~Our models, taking only 4 images as input, have learned to parse the illumination of a quickly changing heightfield surface and use it to predict the long-term (up to 10x the length of initial observation) motion of an object.
    (d) For homogeneously lit flat regions, it is difficult to make decisions, indicated by \PNet's large uncertainty estimates.
    (e) IFS, \TNet and \PTNet correctly interpret the ball's initial angular velocity to predict the future path. 
    \PNet demonstrates the power of anisotropic uncertainty estimation (c, f). It is more certain in the direction of motion than orthogonal to it.
    Note, that NPE and IFS were given the ground truth object positions for the first four frames, and do not have the capability to take images as input.
    }}
    \label{fig:bowlCollision_plate}
\end{figure*}

We also trained our models with two and three balls in the \bowltwoq environment to study the ability of our models to handle object interactions without explicit collision modules. The aforementioned training setups are maintained in these experiments. Quantitatively, \Cref{t:multiple} shows that our models were able to get competitive results \textit{w.~r.~t.} state-only methods containing explicit collision modules, \eg,  NPE. Probabilistic model shows an increase in uncertainty at $T_\text{train}$, which reveals that the task to solve were harder due to the chaotic nature of the system. In addition, angular velocity seems to be very challenging to estimate in this case. Qualitatively, \Cref{fig:bowlCollision_plate} shows that collisions are well handled by our model despite not being explicitly encoded.
\input{tabs/tab_mulballs.tex}

\paragraph{(iii) Ablation study} \revision{To better assess the performance of our model with and without the summation module of }\cref{fig:multiple}\revision{\ we conducted an ablation study. We trained the }\TNet network on both the \bowltwo 2 balls and the \bowltwo \revision{3 balls dataset with and without the extension (with an ablation of the yellow part in }\cref{fig:multiple}). \revision{We first evaluated its performance on long-term predictions. Then we studied how both of these models are handling collisions}.

\input{tabs/ablation_long_term.tex}

\revision{We show in }\cref{t:ablation_lterm} \revision{that the model performs similarly on training regimes (same errors for two balls, 5.2/5.1 and 3.9/3.7 for 3 balls). The module seems to have a clear advantage on long-term predictions where pixel errors are respectively 17.6/16.8 for two balls and 18.2/15.9 for three balls. Angular velocity errors is marginally better without the module, however, both errors remain very close (3.7/4.1 and 4.8/4.9)}.

\revision{Furthermore, we study the impact of our module on collisions. To this end, we created a new dataset extracted from the test data of the multiple balls experiment. In this dataset, we  run a collision detector and clipped the experiment at 10 time steps prior to the first observed collision between the balls.} In \cref{t:ablation_collisions} \revision{we report error numbers at $T=5$ and $T=10$ after the collision. We see that our module enables our pipeline to better handle collisions between objects. }

\input{tabs/ablation_collision.tex}

\ifdefined\realdata
\paragraph{\revisionm{(iv) Real data}}
\revisionm{We also investigate extrapolation on real data using the \textsc{Roll4Real} dataset by} \cite{EhrhardtEtAl:UnsupervisedIntuitivePhysics:ACCV:18}. \revisionm{In our setting, we are only interested in using their unsupervised signal as ground truth position to train our models. We do not address the complex problem of obtaining this signal from unsupervised data. We report the results} in~\cref{t:one-real}, \revisionm{where all models were trained to predict position only. In each scenario, our models were able to handle the transition to real data as opposed to the baselines. For instance for an ellipsoidal bowl} (\bowltwo dataset in \cref{t:one} and \textsc{BowlR1b} in \cref{t:one-real})\revisionm{, errors at \Ttrain } \revisionm{for models trained without angular velocity, went from 1.6} in \cref{t:one} \revisionm{to 5.6} in \cref{t:one-real} \revisionm{for} \PTNet \revisionm{whereas the error went from 3.3 to 26.2 for IFS. The errors at \Text } \revisionm{in this case being generally large ($> 23$) for models trained without angular velocity.} 

\input{tabs/tab_real.tex}
\else
\fi

\input{tabs/tab_interp.tex}

\subsubsection{Interpolation}

So far, we have consider the problem of extrapolating trajectories without any information on the possible final state of the system. We aim here to study the impact of injecting such knowledge in our networks.

In order to do so, in this experiment we concatenate to the first $T_0=4$ input frames the last  observed frame at time $T_\text{final}$ and give the resulting stack as input to the encoder network $h_0=\phi_\text{enc}(\bx_{(-T_0,0]},x_{T_\text{final}})$ to estimate the first state $h_0$. In this setting, the model performs ``interpolation'' as it sees images at the beginning as well as the end of the sequence. The rest of the model works as before with the exception that the first state $h_0$ is decoded in a prediction $(y_0,y_{T_\text{final}}) = \phi_\text{dec}(h_0)$ of both the first and the last position $y_{T_\text{final}}$; in this manner, the loss encourages state $h_0$ to encode information about the last observed frame $x_{T_\text{final}}$.

\cref{t:interp} indicates that the ability of observing an image of the final state enables our models to provide far better estimations. Even in the more complex scenarios with 2 and 3 balls and the heightfield experiments, the errors are significantly lower than for extrapolation. As expected, for InterpNet the highest errors are always found in the middle of the estimate as these points are less predictable from the available information; by contrast, for DispNet the highest errors are at the end. 

Still, we note that harder scenarios result in larger errors even for interpolation, and particularly for colliding balls due to the chaotic nature of this dynamics. This also shows the current limitation of our system in modeling collisions and complex variable environments.

\subsection{Discussion}
In addition to the various results we presented, we discuss our conclusions regarding the main sources of prediction error in the conducted experiments.\\

\noindent\textit{Does training for longer horizons help?}
Training for longer horizons \Ttrain $=40$ in \Cref{t:interp} compared to \Ttrain $=20$ in \mbox{\Cref{t:one}} results in better position estimates %
as expected. When a single end state is also observed (interpolation) the model manages to infer plausible trajectories even though the initial and final states are far apart in time%
. 
\begin{table}[h!]
\centering
\caption{\textbf{Length of supervision.} The \textbf{maximum} position error of \TNet decreases when we add more supervision during training.}
\vspace{0.7em}
\footnotesize
\begin{tabular}{|c|c|c|c|c|c|}
\hline
 Dataset & \multicolumn{2}{c|}{Extrapolation} &  Interpolation \\
 & \Ttrain $=20$ & \Ttrain $=40$ & \Ttrain $=40$ \\
 \hline
\bowloneq & 15.9 & 5.0 & 1.8 \\
\bowltwoq & 16.1 & 3.0 & 1.6 \\
`Heightfield' & 21.6 & 17.9 & 5.2\\
\hline
\end{tabular}
\label{t:horizon}
\end{table}

This motivates us to design more structured representations in the future, which would generalize even better outside the supervised time spans (see \Cref{t:horizon}).\\

\noindent\textit{Can the models handle collisions of multiple objects?}
Adding additional objects to our scenes has appeared to be a challenging task for our models. \revision{If our multiple objects module helped to better handle collisions}\ (see~\cref{t:ablation_collisions}),\revision{ the error increased with the number of objects,} 
which shows that collisions remain difficult to estimate. Promisingly, \IntNet manages to improve performance similarly to the earlier cases, the remaining ambiguity in the middle of the sequences matches the ratios of single object examples ($Error_{T=10}/Error_{T=20}$: $1.0/1.6 \simeq 3.2/4.5 \simeq 3.3/4.5$ in \Cref{t:interp} middle columns).\\

\noindent\textit{Does regression of angular velocity help?} Almost all models benefit from the additional supervision signal coming from the loss on angular velocity, as shown in \Cref{fig:bars_ellipse}(left) and \Cref{fig:bars-heightfields} (left). The objects' texture at these resolutions is difficult to interpret, and the connection between pixel color and rotation around axis is highly non-linear, which encourages us to look for a different representation of rotation in the future to improve our angular prediction errors.\\

\noindent\textit{Are changing environments more difficult?}
The characteristics of the environment also appear to strongly contribute to the final estimation errors. When only following one ball we notice that for simple shapes where the environment parameters can vary along at most 3 dimensions (in the \bowloneq and \bowltwoq cases), the system can obtain nearly perfect estimates in the interpolation experiments. However in the `Heightfield' scenes interaction with the environment is much more difficult to estimate and the maximum errors are larger, even for \IntNet the errors remains substantial.

%% file: tabs/tab_oneball_synth.tex
\setlength{\tabcolsep}{0.4em}
\newcolumntype{L}{S[table-format=-1.2]}
\begin{table*}[h!]
\centering
\caption{\textbf{Long term predictions.}  All of our models (below thick line) observed the $\ninputs=4$ first frames as input. All networks have been trained to predict the $T_\text{train}=20$ first positions, except for the NPEs which were given $\ninputs=4$ states as input and train to predict state at time $\ninputs+1$. We report here results for time $T_\text{train}=20$ and $T_\text{gen}=40$. Unless noted, reported models are trained to predict position and angular velocity. For each time we report on the left average pixel error and root squared $L^2$ angular velocity loss on the right. Perplexity ($\log_e$ values shown in the table) is defined as $2^{-\mathbb{E}[\log_2(p(x))]}$ where $p$ is the estimated posterior distribution. This value is shown in bracket.}
\vspace{0.7em}
\scriptsize
\sisetup{detect-weight=true,detect-inline-weight=math,  table-column-width=3.5em}
\begin{tabular}{|c|C{3em}|*4L|*4L|*4L|*4L|}
\hline
    & & \multicolumn{4}{c|}{\bowlone}  & \multicolumn{4}{c|}{\bowltwo} & \multicolumn{4}{c|}{Heightfield} \\
    Method &  State &
    \multicolumn{4}{c|}{Errors (Perplexity)} &
    \multicolumn{4}{c|}{Errors (Perplexity)} &
    \multicolumn{4}{c|}{Errors (Perplexity)} \\
    & & \multicolumn{2}{c}{\Ttrain} & \multicolumn{2}{c|}{\Text} &
    \multicolumn{2}{c}{\Ttrain} &
    \multicolumn{2}{c|}{\Text} & \multicolumn{2}{c}{\Ttrain} & \multicolumn{2}{c|}{\Text}\\[0.1cm]
    
    &
    & \multicolumn{1}{c}{\pos} 
    & \multicolumn{1}{c}{\av} 
    & \multicolumn{1}{c}{\pos} &\multicolumn{1}{c|}{\av}
    & \multicolumn{1}{c}{\pos} 
    & \multicolumn{1}{c}{\av} 
    & \multicolumn{1}{c}{\pos} &\multicolumn{1}{c|}{\av}& \multicolumn{1}{c}{\pos} 
    & \multicolumn{1}{c}{\av} 
    & \multicolumn{1}{c}{\pos} &\multicolumn{1}{c|}{\av}\\
    
    \hline
    Linear & GT & 39.2 & 7.5 & 127.5
    & 17.9 & 61.9 & 23.3 & 20.1 & 80.0 & 21.3 & 9.4 & 61.9  & 19.3 \\
    \hline
    Quadratic & GT & 164.3 & 18.4 & 120.1 & 861.2 & 11.7 & 14.8 & 93.1 & 70.6 &  26.7 & 27.4 & 126.0  & 122.2 \\
    \hline
    
    NPE \woa & GT & 2.6 & \noentry & 6.0  & \noentry & 3.2 & \noentry & \bfseries 6.1 &  \noentry & 12.0 & \noentry & 38.5  & \noentry \\
    \hline
    NPE & GT & 3.3 &  0.8  & 9.6 &  1.7  & 2.7 & 1.4  & 7.6 & 2.9 &  10.9 & 3.7 & 32.9  & 4.6 \\
    \hline

     \revision{V-LSTM} \woa & Visual & 6.3 & \noentry  & 57.5 & \noentry  & 3.2 & \noentry  & 30.4 & \noentry & 8.8 & \noentry & 26.7  & \noentry \\
    \hline
    \revision{V-LSTM} & Visual & 5.7 & 1.3  & 35.0 & 2.5 & 4.0 & 0.8 & 39.9 & 6.0 & 8.8 & 2.2  & 26.1 & 2.9 \\
    \hline

    IFS \woa & GT & 1.3 & \noentry  & 2.9 & \noentry  & 3.3 & \noentry  & 8.9 & \noentry & 10.4 & \noentry & 27.6  & \noentry \\
    \hline
    IFS & GT & 1.6 & \bfseries 0.3  & \bfseries 2.2 & \bfseries 0.4  & 3.1 & 1.0  & 6.9 & \bfseries 1.4 &  8.7 & 2.5 & 26.1  &  2.8 \\
    \hline
    
    VIN \woa & Visual & 40.4 & \noentry & 37.8 & \noentry & 24.0 & \noentry & 30.2 & \noentry & 42.6 & \noentry  & 42.7 & \noentry\\
    \specialrule{1pt}{0pt}{0pt}
    \PTNet \woa & Visual & \bfseries 1.0 & \noentry & 18.1 & \noentry & 1.6 & \noentry & 24.4 & \noentry & 7.2 & \noentry  & 24.6 & \noentry\\
    \hline
    \PTNet  & Visual & \bfseries 1.0 & 0.4 & 13.8 & 3.0 & \bfseries 1.2 & \bfseries 0.5 & 11.8  & 3.0 &  6.8 & 2.1 & 23.2  & 4.2 \\
    \hline
    
    \TNet \woa & Visual & 3.0 & \noentry & 29.7 & \noentry & 2.5 & \noentry & 20.6 & \noentry &  7.7 & \noentry  & 25.8 & \noentry\\
    \hline
    \TNet & Visual & 3.5 & 1.2 & 15.9 & 4.3 & {\bfseries 2.1} & 1.0 & 16.1 & 4.4  & 7.2 & \bfseries 2.0 & \bfseries 21.6  & 3.3 \\
    \hline
    
    \PNet \woa & Visual & 2.9  & \noentry & 24.2 & \noentry & 2.9 & \noentry & 21.8 & \noentry  & \bfseries 6.4 & \noentry & 22.5 & \noentry\\
               && {\bfseries (4.5)}  &&  {(21.9)}  && {(32.1)} && {(54.0)} && {(\bfseries 9.5 )} && {(12.7)} & \\
    \hline 
    \PNet & Visual & 3.4  & 1.2 & 15.3 & 3.4 & 4.0 & 1.8 & 16.7 &  3.8 & 6.8 & 2.1 & 20.5  & \bfseries 2.7\\
                   && {(4.7)} && {\bfseries ( 9.2)}   && {\bfseries (4.5)} && {\bfseries (9.3)} && {(10.8)} && {(\bfseries 12.3 )} & \\
\hline  
\end{tabular}
\label{t:one}
\end{table*}

%% file: tabs/tab_mulballs.tex
\setlength{\tabcolsep}{0.4em}
\begin{table}[h!]
\caption{\textbf{Multiple balls experiment.} We extend the \bowltwoq setup adding more balls. We show that in this case our networks get comparable performances to state-only methods. Table layout and measures are the same as \cref{t:one} except that $T_ {train}=15$ and $T_\text{gen}=30$.}
\vspace{0.7em}

\centering
\tiny
\sisetup{detect-weight=true,detect-inline-weight=math, table-column-width=2.6em}
\begin{tabular}{|c|C{3em}|*4S[table-column-width=3.5em , table-format=-2.2]|*4S[table-column-width=3.5em ,table-format=-2.2]|}
\hline
    & & \multicolumn{4}{c|}{\bowltwo 2 balls}  & \multicolumn{4}{c|}{\bowltwo 3 balls}   \\
    Method &  States &
    \multicolumn{4}{c|}{Errors (Perplexity)} &
    \multicolumn{4}{c|}{Errors (Perplexity)} \\
    & & \multicolumn{2}{c}{\Ttrain} & \multicolumn{2}{c|}{\Text} & \multicolumn{2}{c}{\Ttrain} & \multicolumn{2}{c|}{\Text}\\[0.1cm]
    && \multicolumn{1}{c}{\pos} 
    & \multicolumn{1}{c}{\av} 
    & \multicolumn{1}{c}{\pos} &\multicolumn{1}{c|}{\av}
    & \multicolumn{1}{c}{\pos} 
    & \multicolumn{1}{c}{\av} 
    & \multicolumn{1}{c}{\pos} &\multicolumn{1}{c|}{\av}\\
    \hline
    NPE & GT & 5.3 & 1.5& 13.4
    & 2.0 & 5.0 & 1.6 & 13.3 & 2.0  \\
    \hline
    \revision{V-LSTM} & Visual & 5.5 & 2.5 & 24.1 & 3.6 & 6.6 & 3.9 & 22.1 & 4.5\\
    \hline
    IFS & GT & 4.1 & \bfseries 1.3 & \bfseries 9.6
    & \bfseries 1.5 &  \bfseries 4.3 & \bfseries 1.5 & \bfseries 10.0 & \bfseries 1.6  \\
    \hline
    \specialrule{1pt}{0pt}{0pt}
    \hline
    \PTNet & Visual & 4.2 & 2.4 & 11.7
    & 2.8 & 5.7 & 4.0 & 15.6 & 4.5\\
    \hline
    \TNet & Visual & \bfseries 3.6 & 2.2 & 16.8
    & 4.1 & 5.1 & 3.7 & 15.9  & 4.9\\
    \hline 
    \PNet & Visual & 5.3 & 2.5 & 19.8
    & 3.6 & 6.5 & 3.9 & 17.1 & 4.1\\
                   && {(7.0)} && {(14.0)}   && {(7.5)} && {(12.6)} &\\
\hline  
\end{tabular}
\label{t:multiple} 
\end{table}

%% file: tabs/ablation_long_term.tex
\setlength{\tabcolsep}{0.4em}
\begin{table}[h!]
\caption{\revision{\textbf{Effect of multiple-ball module on extrapolation.} We study the impact of our multiple objects module of }\cref{fig:multiple}\revision{on the quality of extrapolation in a multiple-ball scenario. We report results for }\TNet \revision{trained with and without the module on long-term prediction tasks. Table layout and measures are the same as} \cref{t:one} \revision{except that $T_ {train}=15$ and $T_\text{gen}=30$.}}

\centering
\scriptsize
\sisetup{detect-weight=true,detect-inline-weight=math, table-column-width=0.8em}
\begin{tabular}{|c|*4S[table-column-width=.8em , table-format=-.2]|*4S[table-column-width=0.8em ,table-format=-.2]|}
\hline
    & \multicolumn{4}{c|}{\bowltwo 2 balls}  & \multicolumn{4}{c|}{\bowltwo 3 balls}   \\
    \cref{fig:multiple} Module &  
    \multicolumn{4}{c|}{Errors} &
    \multicolumn{4}{c|}{Errors} \\
    & \multicolumn{2}{c}{\Ttrain} & \multicolumn{2}{c|}{\Text} & \multicolumn{2}{c}{\Ttrain} & \multicolumn{2}{c|}{\Text}\\[0.1cm]
    &\multicolumn{1}{c}{pix.} 
    & \multicolumn{1}{c}{\av} 
    & \multicolumn{1}{c}{pix.} &\multicolumn{1}{c|}{\av}
    & \multicolumn{1}{c}{pix.} 
    & \multicolumn{1}{c}{\av} 
    & \multicolumn{1}{c}{pix.} &\multicolumn{1}{c|}{\av}\\
    \hline
    $\times$ &  \bfseries 3.6 & \bfseries 2.2 & 17.6
    & \bfseries 3.7 & 5.2 & 3.9 & 18.2  & \bfseries 4.8\\
    \hline 
    \checkmark &  \bfseries 3.6 & \bfseries 2.2 & \bfseries 16.8
    & 4.1 & \bfseries 5.1 & \bfseries 3.7 & \bfseries 15.9  & 4.9\\
    \hline 

\end{tabular}
\label{t:ablation_lterm} 
\end{table}

%% file: tabs/ablation_collision.tex
\setlength{\tabcolsep}{0.4em}
\begin{table}[h!]
\caption{\revision{\textbf{Effect of multiple-ball module on collision estimation.} We study the impact of our multiple objects module of }\cref{fig:multiple} \revision{on collision estimation. All experiments start at $T_0 = T_{\text{first collision}}-10$ for the two multiple balls dataset. We report results for} \TNet \revision{trained with and without the module trained on the extrapolation task in }\cref{ssec:experiments} \revision{with $T_ {train}=15$ and $T_\text{gen}=30$. Table layout and measures are the same as }\cref{t:one}. \revision{We report error at different time $T$ after collision occur}.}

\centering
\scriptsize
\sisetup{detect-weight=true,detect-inline-weight=math, table-column-width=0.8em}
\begin{tabular}{|c|*4S[table-column-width=0.8em , table-format=-.2]|*4S[table-column-width=0.8em ,table-format=-.2]|}
\hline
    & \multicolumn{4}{c|}{\bowltwo 2 balls}  & \multicolumn{4}{c|}{\bowltwo 3 balls}   \\
     \cref{fig:multiple} Module &  
    \multicolumn{4}{c|}{Errors} &
    \multicolumn{4}{c|}{Errors} \\
    & \multicolumn{2}{c}{$T=5$} & \multicolumn{2}{c|}{$T=10$} & \multicolumn{2}{c}{$T=5$} & \multicolumn{2}{c|}{$T=10$}\\[0.1cm]
    &\multicolumn{1}{c}{pix.} 
    & \multicolumn{1}{c}{\av} 
    & \multicolumn{1}{c}{pix.} &\multicolumn{1}{c|}{\av}
    & \multicolumn{1}{c}{pix.} 
    & \multicolumn{1}{c}{\av} 
    & \multicolumn{1}{c}{pix.} &\multicolumn{1}{c|}{\av}\\
    \hline
    $\times$ &  3.3 & 2.9 & 6.6
    & 2.7 & 4.5 & 4.1 & 8.3  & 3.9\\
    \hline 
    \checkmark &  \bfseries 2.6 & \bfseries 2.3 & \bfseries 5.3
    & \bfseries 2.5 & \bfseries 3.9 & \bfseries 3.5 & \bfseries 6.7 & \bfseries 3.7\\
    \hline 

\end{tabular}
\label{t:ablation_collisions} 
\end{table}

%% file: tabs/tab_real.tex
\begin{table}[t!]
\centering
\caption{\textbf{Long term predictions using real data.} \revisionm{All models are trained using the unsupervised tracker output of} \cite{EhrhardtEtAl:UnsupervisedIntuitivePhysics:ACCV:18}, \revisionm{with the same name for every dataset. Reported number are \textbf{pixel errors} for every time. State are the same as}~\cref{t:one}. \revisionm{First three dataset use one ball while last one uses two balls. In all experiment} \Text=$2\times$\Ttrain.}\label{t:one-real}

\vspace{-.7em}
\tiny
\sisetup{detect-weight=true,detect-inline-weight=math, table-column-width=4.0em}
\newcommand{\boldentryy}[2]{%
\multicolumn{1}{S[table-format=-0.1,
                    mode=text, text-rm=\fontseries{b}\selectfont
                   ]#2}{#1}}
\newcolumntype{F}{S[table-format=-0.1]}
\begin{tabular}{|C{3.7em}|*2F|*2F|*2F|*2F|}
\hline
    & \multicolumn{2}{c|}{\textsc{PoolR1b}}
    & \multicolumn{2}{c|}{\textsc{HeightR1b}}
    & \multicolumn{2}{c|}{\textsc{BowlR1b}} 
    & \multicolumn{2}{c|}{\textsc{HeightR2b}} \\
    Method & \multicolumn{2}{c|}{pix. err, $T_{train}= 15$} & \multicolumn{2}{c|}{pix. err, $T_{train}= 20$} & \multicolumn{2}{c|}{pix. err, $T_{train}= 20$} & \multicolumn{2}{c|}{pix. err, $T_{train}= 15$} 
    \\
    & \multicolumn{1}{c}{\Ttrain} &
    \multicolumn{1}{c|}{\Text} & \multicolumn{1}{c}{\Ttrain} &
    \multicolumn{1}{c|}{\Text} & \multicolumn{1}{c}{\Ttrain} &
    \multicolumn{1}{c|}{\Text}&
    \multicolumn{1}{c}{\Ttrain} &
    \multicolumn{1}{c|}{\Text}\\[0.1cm]
\hline
    V-LSTM & 6.5 & 30.4 & 6.1 & 31.3 & 10.9 & 58.8 & 19.0 & 38.2\\
\hline
    IFS &  26.0 & 37.5 &   48.0  & 58.1  & 26.2  & 39.1  & 15.6& 26.6\\
\hline
    VIN &  50.9 & 40.8 & 40.2  & 47.3  & 33.9   & 33.0 & 45.9 & 39.8\\
\specialrule{1pt}{0pt}{0pt}
    \PTNet & 4.6 & 21.4 & \boldentryy{5.6}{} & 29.0  &  \boldentryy{5.6}{} & 23.0 & \boldentryy{5.4}{}& \boldentryy{12.5}{|}\\
\hline
    \TNet &  \boldentryy{3.8}{} & 23.6 & \boldentryy{5.6}{}  & \boldentryy{28.5}{|}&  6.5  & \boldentryy{22.6}{|} & 6.2 & 15.4\\
\hline
 \PNet & 4.7{(6.)} &  \boldentryy{16.3\textsc{{(11.)}}}{|}  & 5.7{(6.)} & 30.0{(22.)}   &  6.8{(7.)} &  23.5{(14.)} & 6.8{(8.)} &  16.9{(12.)}\\
\hline
\end{tabular}
\end{table}

%% file: tabs/tab_interp.tex
\newcolumntype{K}{S[table-format=0.1, table-column-width=2.0em]}
\begin{table*}[t!]
\centering
\caption{\textbf{Extrapolation vs Interpolation.} We constructed \IntNet as an extension of \TNet, where in addition to the concatenation of the first $\ninputs=4$ frames, also the last frame at $T_{final}$ is provided to the model as inputs. All networks have been trained to predict the $T_\text{train}:=T_{final}$ positions. As expected, \IntNet learned to predict the positions at $T_{final}$ by relying on the features extracted from the last input image. We report the \textbf{pixel errors} at different times along the sequences. $T_{final}$ is the last value shown for every experiment.}
\vspace{0.7em}
\footnotesize
\sisetup{detect-weight=true,detect-inline-weight=math, table-column-width=2.0em}
\begin{tabular}{|C{4em}|*4K|*4K|*3K|*3K|*4K|}
\hline
    & \multicolumn{4}{c|}{\bowlone}  & \multicolumn{4}{c|}{\bowltwo 1 ball} & \multicolumn{3}{c|}{\bowltwo 2 balls} & \multicolumn{3}{c|}{\bowltwo 3 balls}& \multicolumn{4}{c|}{Heightfield} \\ 
    Method & \multicolumn{4}{c|}{pixel error, $T_\text{train}=40$} &
    \multicolumn{4}{c|}{pixel error, $T_\text{train}=40$} & \multicolumn{3}{c|}{pixel error, $T_\text{train}=30$} 
    & \multicolumn{3}{c|}{pixel error, $T_\text{train}=30$} 
    & \multicolumn{4}{c|}{pixel error, $T_\text{train}=40$} \\
    & {T=10} & {20} & {30} & {40}
    & {T=10} & {20} & {30} & {40}
    & {T=10} & {20} & {30}
    & {T=10} & {20} & {30} &
    {T=10} & {20} & {30} & {40}\\
    \hline 
    \TNet & 2.2 & 3.6 & 3.9 & 5.0 & 1.4 & 2.4 & 2.7 & 3.0
    & 2.8 & 5.8 & 8.7 & 3.2 & 8.1 & 12.0 & 3.6 & 7.9 & 12.9 & 17.9  \\    
    \hline 
    \IntNet & 1.4 & 1.8 & 1.6 & 1.0 & 1.0 & 1.6 & 1.3 & 0.6 &  3.2 & 4.5 & 3.1 & 3.3 & 4.5 & 2.1 & 2.5 & 5.2 & 5.1 & 1.6 \\
\hline    
\end{tabular}
\label{t:interp}
\end{table*}

%% file: conclusions.tex
\section{Conclusions}\label{s:conclusions}

In this paper, we studied the possibility of abstracting knowledge of physics using a single neural network with a recurrent architecture to model long term predictions with a changing environment. We compared our model to various baselines on the non-trivial motion of  ball(s) rolling on a surfaces with different possible shapes (\eg ellipsoidal bowls or randomized heightfields) \revisionm{on both synthetic and real data}. %
\revisionm{Closer to some approaches}, we do \emph{not} integrate physical quantities but implicitly encode the states in a feature vector that we can propagate through time.

\revisionm{However, }we demonstrated a significant difference compared to existing networks \revisionm{using implicit state encoding}, namely the ability to account for complex variable environments. The latter leverage a distributed representation of the system state which, at the same time, is still able to model concentrated object interactions such as collisions. 

Our experiments on synthetic simulations \revisionm{also} indicate that our networks can predict mechanical phenomena more accurately than networks that build on hand-crafted physically-grounded representations of the system state. This means that our approach can both infer automatically an internal representation of these phenomena and  work with visual inputs in order to initialize such a representation and use it for extrapolation. Our models can also estimate a distribution over physical measurements such as position to account for uncertainty in the predictions.

While keeping the same architecture, we further demonstrate that it is possible to remove ambiguity by showing the network an image of the final state of the system, performing interpolation. However, in this case the internal state propagation mechanism is still limited by its ability to make accurate long term predictions outside temporal spans observed during training.

In the future, we aim at increasing  the robustness and generalization capabilities of our models by enforcing more explicitly temporal and spatial invariance (as physical laws are constant and homogeneous). 
\ifdefined\realdata
\else
\fi
\revision{Finally, we plan to work on the generalization abilities of our multiple objects pipeline to handle various object shapes and remove the limitation of having to known the number of objects in advance.}